\documentclass[italian,french]{amsart}
\usepackage[T1]{fontenc}
\usepackage[latin9]{inputenc}
\usepackage{url}
\usepackage{amsbsy}
\usepackage{amsthm}
\usepackage{amssymb}
\usepackage{graphicx}

\makeatletter

\providecommand{\tabularnewline}{\\}

\numberwithin{equation}{section}
\numberwithin{figure}{section}
\newenvironment{lyxcode}
{\par\begin{list}{}{
\setlength{\rightmargin}{\leftmargin}
\setlength{\listparindent}{0pt}
\raggedright
\setlength{\itemsep}{0pt}
\setlength{\parsep}{0pt}
\normalfont\ttfamily}%
 \item[]}
{\end{list}}

\usepackage{etoolbox}

\patchcmd{\@maketitle}
  {\ifx\@empty\@dedicatory}
  {\smallskip
    \begin{center}
    \footnotesize
    \begin{tabular}{c}
     Professeur, IMAG, Univ Montpellier, CNRS \\
    \end{tabular}
  \end{center}
  \ifx\@empty\@dedicatory}
  {}{}

\makeatother

\usepackage{babel}
\makeatletter
\addto\extrasfrench{%
   \providecommand{\fg}{\ifdim\lastskip>\z@\unskip\fi~\frqq}%
}

\makeatother
\begin{document}

\title{Critères de qualité d'un classifieur généraliste}

\author{Gilles R. Ducharme}

\keywords{Classifier, Logistic Regression, Random Forest, Neural Network, Predictor,
Supervised Learning, Support Vector Machine.}
\begin{abstract}
This paper considers the problem of choosing a good classifier. For each problem there exist  an optimal classifier, but none are optimal, regarding the error rate, in all cases. Because there exists a large number of classifiers, a user would rather prefer an all-purpose classifier that is easy to adjust, in the hope that it will do almost as good as the optimal.
In this paper we establish a list of criteria that a good generalist classifier should satisfy .
We first discuss data analytic, these criteria are presented. Six among the most popular classifiers are selected and scored according to these criteria. Tables allow to easily appreciate the relative values of each. In the end, random forests turn out to be the best classifiers.

 - - - - -

Cet article considère le problème de choisir un bon classifieur. Pour chaque contexte  il existe un classifieur optimal selon le critère du taux d'erreur, mais aucun n'est optimal dans tous les cas. Comme il existe de nombreux classifieurs, lÕutilisateur préférera souvent choisir un classifieur généraliste, dont l'ajustement et l'exploitation sont à sa portée, en espérant que celui-ci fait presque aussi bien que l'optimal. Cet article établit une liste de critères que devrait rencontrer un bon classifieur généraliste, destiné à être ajusté et utilisé avec un minimum d'intervention humaine. Après avoir introduit l'analytique des données, ces critères sont présentés et commentés. Puis un sous-ensemble de six classifieurs est choisi parmi les plus populaires et des scores leur sont attribués en regard de ces critères. Des tables permettent d'apprécier les résultats et facilitent le choix d'un bon classifieur. Le classifieur qui ressort de cet exercice avec les meilleurs scores est la forêt aléatoire et ses variantes.léatoire (random forest) et
ses variantes.
\end{abstract}

\maketitle

\section{Introduction}

La data-science\footnote{L'Université de Montpellier (à l'époque, Université de Montpellier
2) est le ``berceau'' de la data science. En effet, ce terme est
apparu lors des travaux de la deuxième rencontre des sociétés françaises
et japonaises de statistique, tenue à l'été 1992 à Montpellier, dont
les \emph{proceedings} ont été publiés en 1995 (Escoufier et al.,
1995). Une version plus complète de la genèse du terme apparaît dans
Ohsumi (2000).} est devenue dans bien des sphères d'activité un incontournable facteur
de développement. Elles émanent de la convergence entre les mathématiques
et l'informatique, plus précisément entre la statistique et le \emph{machine
learning}, et visent à exploiter les immenses bases de données qui
se constituent grâce au phénomène \emph{big data}. C'est souvent par
le biais des data-scientists que l'intelligence artificielle, dont
le \emph{machine learning} fait partie, s'introduit dans les entreprises.
Un de ses outils les plus utilisés est l'analytique des données et
notamment l'analytique prédictif. La prédiction du futur étant un
des déterminants de la survie, l'analytique prédictif s'est répandu
dans de nombreuses disciplines et champs d'activité, allant du marketing
à la médecine, en passant par le management.

L'analytique prédictif utilise un même outil, le \emph{classifieur},
pour résoudre une vaste gamme de problèmes. Un de ces problèmes est
celui de déterminer, à partir de ses \emph{caractéristiques,} l'état
d'une situation cachée. On réfère à ce contexte comme étant un problème
de classification. Un exemple prototype est le système de reconnaissance
faciale sur certains téléphones. \emph{A priori,} le téléphone ``ignore''
si un utilisateur est autorisé ou non, à accéder à son contenu. Les
caractéristiques de l'image faciale (écartement des yeux, forme des
oreilles, zones sombres, etc.) d'un utilisateur potentiel sont soumises
à un classifieur qui doit déterminer si cet utilisateur fait partie,
ou non, de la liste des usagers autorisés. Le but de l'analytique
prédictif est ici de construire un classifieur qui présentera le taux
d'erreur le plus faible possible. On retrouve aujourd'hui de nombreuses
variantes de ce type de classifieur, en reconnaissance d'images, de
caractères, de sons, etc. L'utilisation croissante de ces classifieurs
dans les objets de la vie courante témoigne de l'efficacité de l'analytique
prédictif dans de tels contextes de classification. Coubray (2017)
fait état de deux applications récentes dans le domaine de la reconnaissance
vocale en marketing.

Un autre problème traité par l'analytique prédictif est celui de la
prédiction. Un exemple prototype est le \emph{credit scoring} où,
à partir de caractéristiques (salaire, nombre de personnes à charge,
crédits en cours, etc.) associées à d'éventuels emprunteurs, une banque
veut déterminer si un client se révèlera à terme un ``bon'' ou un
``mauvais'' payeur. Ici aussi, la prédiction est faite par un classifieur
(on utilise alors parfois le terme de \emph{prédicteur}) : les caractéristiques
du client potentiel sont soumises au classifieur qui détermine l'état
futur, ``bon'' ou ``mauvais'', avec un taux d'erreur le plus faible
possible. L'usage très répandu de ces classifieurs dans le domaine
bancaire, commercial, marketing, etc., témoigne de leur efficacité
dans ce contexte de prédiction. La même chronique de Coubray (2017)
décrit une application pour la réduction du taux de \emph{churn}. 

Un premier axe permettant de distinguer ces deux contextes est la
temporalité de la réalité que l'on cherche à découvrir. Dans l'exemple
prototype du contexte de classification (le téléphone), cette réalité
existe, mais est cachée. Dans celui du contexte de prédiction (le
\emph{credit scoring}), la réalité se révèlera dans le futur. Un second
axe est la nature des caractéristiques dont on dispose pour faire
la classification ou la prédiction. Dans l'exemple prototype du téléphone,
ces caractéristiques sont les éléments de l'image faciale qui, une
fois enregistrés sur un support, demeurent statiques. Il y a peu ou
pas de hasard (dans le sens de la théorie des probabilités) en jeu.
Dans celui du contexte de prédiction, les caractéristiques se rapportent
à des individus. Or les humains changent, évoluent, s'adaptent; leurs
caractéristiques feront de même. En outre, les humains diffèrent entre
eux et le hasard, ou la casualité joue un rôle important dans ces
changements et ces différences. On retrouve toute une gamme de problèmes
de classification et de prédiction se situant quelque part sur le
plan formé de ces deux axes : temporalité et casualité.

Chaque classifieur est articulé autour d'un concept mathématique et
incarné par un algorithme informatique. C'est la phase d\emph{'invention}
du classifieur. À la sortie de cette phase, qui est en général hors
de la portée des utilisateurs, on peut imaginer le classifieur comme
étant une ``boîte noire'' ayant des manettes, curseurs ou boutons,
bref des \emph{hyperparamètres}\footnote{Pour faire simple, ce terme regroupe ici toutes les quantités (fonctions,
variables d'ajustement, structures mathématiques, etc.) qui doivent
être sélectionnées ou déterminées pour qu'un classifieur soit opérationnel.} qu'il faut maintenant ajuster, ou positionner aux bons endroits,
pour résoudre au mieux le problème considéré (p. ex. faire en sorte
que le taux d'erreur est sous un seuil donné en \emph{credit scoring}).
Cette étape est la phase \emph{d'ajustement} du classifieur et elle
s'appuie en général sur les informations (cognitives ou empiriques)
dont on dispose sur le problème considéré. Une fois ces hyperparamètres
ajustés, on présente des caractéristiques (images, dossiers clients)
à la boîte noire qui en ressort une classification ou une prédiction.
C'est la phase \emph{d'exploitation }du classifieur.

Depuis le travail fondateur de Fisher (1936), de nombreux concepts
ont été imaginés et des algorithmes les matérialisant ont été mis
au point. Lim, Loh \& Shih (2000) en compilent 33 et plusieurs autres,
à la complexité grandissante, se sont rajoutés depuis. Au moment d'écrire
ces lignes, le package caret (Kuhn, 2008) du logiciel R peut traiter
238 classifieurs\footnote{Parmi ceux-ci, on retrouve cependant de nombreuses variantes du même
classifieur, de sorte qu'on peut estimer à environ une centaine le
nombre de classifieurs ``différents'' que ce package peut reconnaître.}. Ceci témoigne de l'inventivité humaine, mais comme le veut l'adage
: ``trop, c'est comme pas assez''. Et maintenant, face à un problème
de classification donné et à cette pléthore de possibilités, l'utilisateur
se heurte à la difficile question : ``\emph{Quel classifieur dois-je
utiliser pour mon problème }?''

Cette question est consternante pour l'utilisateur qui, s'il fait
un mauvais choix, risque d'utiliser un classifieur menant à des décisions
désastreuses. Elle est surtout mal posée, car aucun critère précis,
à part l'objectif flou de faire ``le mieux possible'', ne se dégage
\emph{a priori} pour y répondre intelligemment. Mais cette question
est néanmoins fondamentale. Et c'est le but du présent travail que
de construire une réponse raisonnée en développant une liste de critères
et en notant, selon ceux-ci, différents classifieurs.

Un critère raisonnable est celui du taux d'erreur du classifieur,
que l'on voudrait le plus faible possible. Ce taux d'erreur mesure
l'écart moyen entre la sortie du classifieur et la réalité. C'est
un critère \emph{accuracy-oriented} et, de ce fait, bien adapté aux
méthodes mathématiques. De nombreux résultats, certains très poussés,
ont été obtenus à son sujet. En particulier, un théorème annonce que,
pour certains types de problèmes, il existe un classifieur, le classifieur
de Bayes, dont le taux d'erreur est minimal entre tous. C'est un joli
résultat mathématique, mais inutile en pratique, car sa détermination
requiert des informations qui ne sont jamais disponibles. En revanche,
il crée une cible irrésistible pour des concepteurs-théoriciens :
la recherche de classifieurs qui ``à la limite'' (et les contextes
\emph{big data} en sont proches) donneront le même taux d'erreur que
le classifieur de Bayes. Ceci pousse à la création de classifieurs
de plus en plus sophistiqués, comprenant de plus en plus d'hyperparamètres
dont l'ajustement nécessite de plus en plus de doigté, mais dont le
maniement exige une expertise s'éloignant de la portée non seulement
de l'utilisateur moyen, mais aussi de celle des concepteurs de méthodologies
concurrentes ! Cette dynamique est encouragée par des compétitions
aux prix financièrement alléchants (Netflix) et aux résultats largement
médiatisés (Kaggle). Elle répond aux impératifs des contextes de classification
pointus que l'on retrouve, notamment, dans les applications du \emph{machine
learning} (reconnaissance d'images, de sons, d'anomalies, etc.) où
la précision brute du classifieur est tout ce qui importe \footnote{Le choix d'une mesure de cette précision brute d'un classifieur n'est
pas une entreprise aisée. Récemment, Hand \& Anagnostopoulos (2013)
ont trouvé une faille fondamentale dans la très populaire métrique
\emph{AUC} (pour \emph{Area Under the Curve}), pourtant utilisée depuis
plus de 40 années, qui la discrédite quasi totalement pour la comparaison
de classifieurs. Le problème se répercute sur d'autres métriques associées,
comme le coefficient de Gini et l'\emph{AUCH. }Voir Hand (2012) pour
une liste de telles métriques avec leurs avantages et inconvénients.}. Dans la suite, nous ne considérons pas plus avant ces contextes
pointus de classification d'images et de ces autres situations statiques
évoquées quand on entend le terme \emph{machine learning}. Nous nous
concentrons plutôt sur les applications en marketing, en santé publique
et en d'autres contextes impliquant des organismes vivants. Aussi,
par opposition à ces problèmes pointus traités en \emph{machine learning}
où le hasard est quasi inexistant, nous allons référer à ces contextes
par le terme de \emph{statistical learning} car les propriétés des
classifieurs découlent des comportements statistiques qui contrôlent
la dynamicité et la casualité des données. 

Pour ces derniers cas, un autre théorème, appelé le \emph{no-free-lunch
theorem} (Wolpert, 1996), annonce qu'il n'existe pas de classifieurs
dominant (toujours en ce qui concerne le taux d'erreur) tous les autres
pour tous les problèmes. Ce résultat crée une nouvelle cible irrésistible
pour des concepteurs adeptes de compromis qui vont chercher le classifieur
ayant le plus faible taux d'erreur possible pour le plus grand nombre
de problèmes possibles. Toutes sortes de classifieurs ont été mis
au point dans cet esprit, notamment les classifieurs d'ensembles et
les méta classifieurs (Dixit, 2017). Mais cette recherche du meilleur
compromis mène aussi à l'élaboration de classifieurs dont, pour la
plupart, le maniement exige un doigté et une expertise de très haut
niveau. 

Au fil des années, ces nombreux classifieurs ont été appliqués à de
nombreux problèmes de la vie courante, alors que plusieurs études
comparatives en laboratoire étaient menées (voir Hand, 2001, pour
une bibliographie très fournie, mais déjà surannée). Ceci a permis
de jeter une certaine lumière sur les forces et les faiblesses de
chacun de ces classifieurs face à un type de problème donné, un savoir
nécessaire pour que l'utilisateur puisse faire un choix éclairé dans
l'offre pléthorique. Mais l'accumulation de connaissances qui en est
résultée est un corpus de plus en plus difficile à s'approprier. Ceci
a mené Hand (2006) à écrire :\medskip{}

\begin{quote}
\emph{Of what good is a tool so highly sophisticated that it can be
used effectively only after years of practice and experience ?}

\medskip{}
\end{quote}
Gayler (2006) a atteint un seuil de clairvoyance élevé lorsqu'il a
constaté (dans sa terminologie, \emph{model }réfère à un classifieur): 

\medskip{}

\begin{quote}
\emph{in the limit (and the hands of a skilled modeler), every modeling
technique should end up in agreement because they are all approximating
the same data.}

\medskip{}
\end{quote}
Puis plus loin :

\medskip{}

\begin{quote}
.\emph{.. it might be more useful to look at the effort required of
the modeler to achieve a given goodness of fit and other properties
of the model that are of operational relevance.}

\medskip{}
\end{quote}
On peut extraire de ces remarques deux enseignements. Le premier,
plutôt rassurant, est que pour les problèmes adaptés au \emph{statistical
learning} et dans les mains d'un expert qui sait bien ajuster les
hyperparamètres, la plupart des classifieurs modernes se valent à
peu près en ce qui concerne le taux d'erreur. Ainsi, la question de
choisir un classifieur \emph{accurate} peut être résolue par une pédagogie
adéquate et beaucoup d'expérience. Le corollaire cependant, introduisant
le deuxième enseignement, est qu'il semble souhaitable d'ajouter aux
critères \emph{accuracy-oriented} d'un classifieur de nouveaux critères,
peut-être plus importants, quantifiant leur accessibilité pédagogique
et leur \emph{operational relevance}. Tout ceci est joliment énoncé
par Jamain \& Hand (2008) :

\medskip{}

\begin{quote}
\emph{..the ``performance of a method'' is not a well-defined term.
In general, how well a method performs is not merely a property of
the method, but is the result of an interaction of the method with
the type of data, the background and experience of the user, the implementation
of the method and the use to which the classification method will
be put.}

\medskip{}
\end{quote}
Cette défocalisation sur les critères \emph{accuracy-oriented} s'exerçait
depuis un certain temps déjà. En \emph{credit scoring} par exemple,
une forte concurrence peut mener une banque à accepter un taux d'erreur
plus élevé si, en échange, elle rencontre les normes légales de transparence
ou si elle peut augmenter ses parts de marché sur d'autres produits.
Plus généralement, quand des décisions sont prises dans un climat
d'incertitude, des risques apparaissent qui doivent être gérés en
fonction de la gravité des conséquences. Les humains sont ainsi faits
qu'en général, ils préfèrent un risque faible associé à une conséquence
limitée (accident de voiture) à un risque infinitésimal, mais pouvant
entrainer des conséquences catastrophiques (crash d'avion). C'est
une des tâches du \emph{statistical learning} que de fournir les supports
intellectuels nécessaires à la prise de décisions dans des ambiances
où la volatilité est largement le fait du hasard. 

Ainsi, les remarques de Gayler (2006) et Jamain et Hand (2008) suggèrent
l'établissement de critères :

\bigskip{}

\begin{itemize}
\item \emph{User-oriented} : comme l'expertise nécessaire pour l'ajustement
(les hyperparamètres) d'un classifieur et pour comprendre et interpréter
son fonctionnement ;
\item \emph{Problem-oriented }: concernant l'usage qui sera fait des sorties
du classifieur.
\end{itemize}
\begin{lyxcode}
\medskip{}
\end{lyxcode}
En ce moment, une intense activité de recherche s'exerce autour de
certains critères tombant dans ces catégories. En particulier, de
nombreuses recherches en cours, notamment dans le \emph{Goodle Brain
Team} (Dean, 2018), portent sur l'automatisation de l'ajustement de
classifieurs. Mais ces recherches portent essentiellement sur les
très populaires classifieurs de type ``réseaux de neurones artificiels''
(voir Section \ref{subsec:Reseaux-de-neurones}) typiques du \emph{machine
learning}. Pour les classifieurs du \emph{statistical learning}, nous
avons trouvé assez peu de travaux sérieux proposant des critères \emph{user-}
et \emph{problem-oriented}, à l'exception notable de Hand (2012) qui
donne une courte liste de quelques critères de type \emph{problem-based}
(selon sa terminologie) dont certains se retrouvent d'ailleurs plus
loin dans ce travail. 

Ainsi, le but du présent travail est de développer une liste plus
complète de critères \emph{user-} et \emph{problem-oriented} permettant
à l'utilisateur de faire le choix éclairé d'un classifieur parmi ceux
qui sont utilisés en \emph{statistical learning}. Pour montrer l'utilité
de cette liste, nous appliquons ces critères à six classifieurs parmi
les plus populaires en \emph{statistical learning}. Des scores sont
donnés, ce qui permet de noter ces classifieurs afin de dégager les
meilleurs. Cette liste de critères n'a pas la prétention de l'exhaustivité.
Elle vise à aider un utilisateur voulant répondre à la question ``\emph{Quel
classifieur dois-je utiliser pour mon problème }?''. Elle reflète
les connaissances et l'expérience professionnelle de l'auteur qui
s'est construite au cours de plus de 35 années dans ses rôles de professeur
et de chercheur universitaire en data science, mais aussi de consultant
et d'expert dans de nombreux domaines d'application allant du marketing
à la biostatistique en passant, entre autres, par la gestion et sécurité
numérique. L'auteur espère aussi qu'elle peut être utile à des utilisateurs
tentant de discerner la part de vérité dans les discours ambiants
tentant d'exploiter la médiatisation à outrance des vertus de l'intelligence
artificielle. Il espère enfin qu'elle favorise l'émergence d'un écosystème
de concepts et d'algorithmes, s'arrimant aux sorties d'un classifieur
et offrant des fonctionnalités permettant d'atteindre des objectifs
au-delà de l'analytique prédictif, comme ceux que l'on voit émerger
dans le domaine de l'analytique prescriptif dont nous discutons brièvement
à la Section \ref{subsec:Les-types-d'analytique}. 

La Section 2 présente sommairement les types d'analytique et le contexte
dans lequel se déroule en général un analytique prédictif. La Section
3 développe la liste de critères et les répartit en trois sous-listes
à l'importance différente. La Section 4 présente brièvement les six
classifieurs qui seront par la suite notés. La Section 5 donne les
tableaux des scores, ce qui permet de dégager les forces et les faiblesses
de chacun. La forêt aléatoire de Breiman (2001, 2004) se dégage nettement
selon ces critères. La Section 6 présente les conclusions finales. 

\section{Analytique des données et classifieurs}

On retrouve des applications de l'analytique des données dans virtuellement
toutes les sphères d'activité. Outre la santé et les sciences, l'analytique
des données est très utilisé dans le domaine industriel et commercial,
comme les services de vente, de marketing, d'approvisionnement, de
gestion, de finance, de risque, de la règlementation, etc. Le phénomène
\emph{big data} a mené à l'accumulation massive de données facilement
accessibles. Celles-ci, couplées à l'exploitation de concepts statistiques
et d'algorithmes informatiques qui en dégagent les structures fines
et cachées, sont un des atouts importants de la compétitivité des
entreprises modernes. La problématique consistant à mettre ces éléments
en oeuvre pour piloter des stratégies favorisant le développement
est un des enjeux capitaux du monde contemporain.

\subsection{Les types d'analytique\label{subsec:Les-types-d'analytique}}

On distingue généralement trois types d'analytique. L'analytique \emph{descriptif}
se concentre sur le passé. Il utilise les données disponibles (données
historiques) pour fournir des réponses à la question \emph{``Que
s'est-il passé ?''} Il produit des graphiques et statistiques (moyenne,
écart-type, etc.) pour comprendre les raisons d'un comportement, d'une
évolution, d'une situation. Il permet l'élaboration d'un \emph{storytelling}.

L'analytique \emph{prédictif} vise à répondre à la question \emph{``Que
va-t-il se passer ?''} Comme il a été dit dans l'Introduction, il
exploite les données au travers de classifieurs. Il donne des prévisions
de ce qu'est la réalité ou des prédictions de cette réalité.

Enfin, l'analytique \emph{prescriptif} (le terme est dû, semble-t-il,
à Lustig et al., 2010) est la dernière frontière de l'analytique.
Il vise à répondre à la question \emph{``Comment améliorer ce qui
devrait se passer ?''}. En utilisant des éléments de recherche opérationnelle
et de statistique, il cherche à définir les actions à entreprendre,
les comportements à modifier, les incitatifs à mettre en place, etc.
dans le but de mieux monétiser la sortie des classifieurs. Ainsi,
le passage du prédictif au prescriptif s'accompagne d'une montée en
puissance des interventions de data scientists. 

La frontière entre l'analytique prédictif et prescriptif est floue
: de façon un peu réductrice, on dit parfois que l\textquoteright analytique
prescriptif est la continuité de l'analytique prédictif dans une optique
opérationnelle. Gayler (2006), avec le terme \emph{operational relevance},
avait déjà pressenti son importance. 

\subsection{L'analytique prédictif }

La phase d\emph{'invention} des classifieurs de l'analytique prédictif
relève du \emph{statistical} et du \emph{machine learning}. Avec le
temps, une fertilisation croisée de principes, de notions et de terminologies
s'est produite, confondant la frontière entre les deux \emph{learning}.
En particulier, il s'est installé une dualité de termes et jargons
utilisés selon qu'on est venu à la data science depuis les mathématiques
ou l'informatique. La confusion semble parfois entretenue pour exploiter
l'effervescence médiatique entourant l'intelligence artificielle.
Dans la suite de ce travail, on a plutôt utilisé la terminologie statisticienne,
d'une part parce que l'auteur en est issu et d'autre part, parce qu'elle
a été polie et durcie par l'usure du temps. Elle n'est pas aussi imagée
que le jargon informatique, mais issue des mathématiques, elle en
a hérité la rigueur, la précision et la cohérence. 

\subsection{Ajustement de classifieur}

Après avoir été \emph{inventé} et avant d'être \emph{exploité} en
production, un classifieur doit être \emph{ajusté} au problème considéré.
Il est commode d'imaginer que cet ajustement nécessite une quantité
d'information spécifique au problème, que l'on note $\mathcal{I}$.
Cette information provient essentiellement de deux sources. La première
est l'ensemble des éléments cognitifs (théories, considérations heuristiques,
connaissances du domaine applicatif, expérience empirique ou de terrain
concernant le comportement du classifieur, etc.) accumulés, que l'on
note $\mathcal{C}$. Elle est matérialisée entre autres par la valeur
de constantes, de variables, de fonctions et par des modèles mathématiques
qui décrivent les grandes lignes du comportement du phénomène étudié.
La deuxième source est un ensemble de données $\mathcal{D}$. Pour
faire simple, on peut imaginer que l'information nécessaire à l'ajustement
d'un classifieur doit satisfaire la relation :
\begin{align*}
\mathcal{I} & \leq\mathcal{C}+\mathcal{D}.
\end{align*}

Si $\mathcal{C}$ est riche, on a besoin de peu de $\mathcal{D}$
(dans certains cas, aucune donnée n'est nécessaire) pour atteindre
le niveau $\mathcal{I}$ nécessaire à l'ajustement du classifieur.
Sinon, il faut compenser par de grandes quantités de $\mathcal{D}$.
Signalons l'asymétrie d'importance entre $\mathcal{C}$ et $\mathcal{D}$
: $\mathcal{C}$ est beaucoup plus informatif que $\mathcal{D}$.
Par exemple, une formule comme $E=mc^{2}$ vaut incommensurablement
plus que n'importe quel amas de données. Ainsi, avant d'ajuster un
classifieur, il importe de recenser les éléments cognitifs dont on
dispose. Il faut discréditer un discours ambiant en data science qui
suggère que l'on puisse faire l'économie de cet effort intellectuel
en le remplaçant par la magie d'algorithmes automatisant cet ajustement.
Dans bien des cas, ceux-ci ne donnent que de premières approximations,
qui demandent à être affinées à la lumière de $\mathcal{C}$.

Lorsque $\mathcal{C}$ est riche, le modèle mathématique du classifieur,
issu de sa conceptualisation, s'écrit sous la forme des fonctions,
ou de règles, mathématiques où tout est connu sauf la valeur d'un
\emph{paramètre} noté $\theta$\emph{}\footnote{En réalité, il existe toujours de plusieurs paramètres. Mais ici,
pour fluidifier le discours, on fait comme s'il n'en existe qu'un
seul.}, une quantité conceptuellement différente d'un hyperparamètre car
elle ne doit pas être choisie, ou sélectionnée, mais plutôt \emph{estimée}
à partir de $\mathcal{D}$. Le classifieur est alors dit \emph{paramétrique}
et en général on a besoin d'un $\mathcal{D}$ relativement petit pour
connaître la valeur approximative de $\theta$. Les classifieurs paramétriques
sont utiles quand $\mathcal{D}$ est pauvre et leur phase d'ajustement
est facile, car ils ne comportent en général que peu ou pas d'hyperparamètres.
La biostatistique et le \emph{credit scoring} sont des domaines où
les données coûtent cher ; les succès incontestables des classifieurs
paramétriques dans ces domaines attestent de leur utilité quand les
données sont rares\footnote{Il convient toutefois de noter qu'un problème de classification ne
``marche pas'' toujours. Un échec de classification veut dire que
le classifieur ne fait guère mieux qu'un choix au hasard. Ces échecs
sont parfois le fait d'un ajustement incorrect, ce qu'un expert arrive
facilement à détecter et à corriger, mais plus souvent d'une pauvreté
d'information discriminante dans $\mathcal{D}$. On peut alors tenter
``d'enrichir'' les données, mais ceci est un exercice qui dépasse
le cadre du présent travail. }. Un avantage supplémentaire est que le paramètre ``parle'' : la
valeur de $\theta$ permet de comprendre comment opère le classifieur
(c.f. celui de la Section \ref{subsec:R=0000E9gression-logistique-(RLog)})
et d'interpréter son fonctionnement, lequel est garant de la confiance
qu'on peut lui accorder.

Il existe de nombreux cas, notamment dans les applications industrielles
et commerciales, où $\mathcal{C}$ est pauvre. Il faut alors compenser
en exploitant des données abondantes (\emph{big data}) pour atteindre
le niveau $\mathcal{I}$. Ces classifieurs sont appelés (par opposition)
\emph{non paramétriques} et font intervenir de façon plus critique
des hyperparamètres.

\subsection{Les données\label{subsec:Les donn=0000E9es}}

\begin{figure}
\begin{centering}
\includegraphics[scale=0.8]{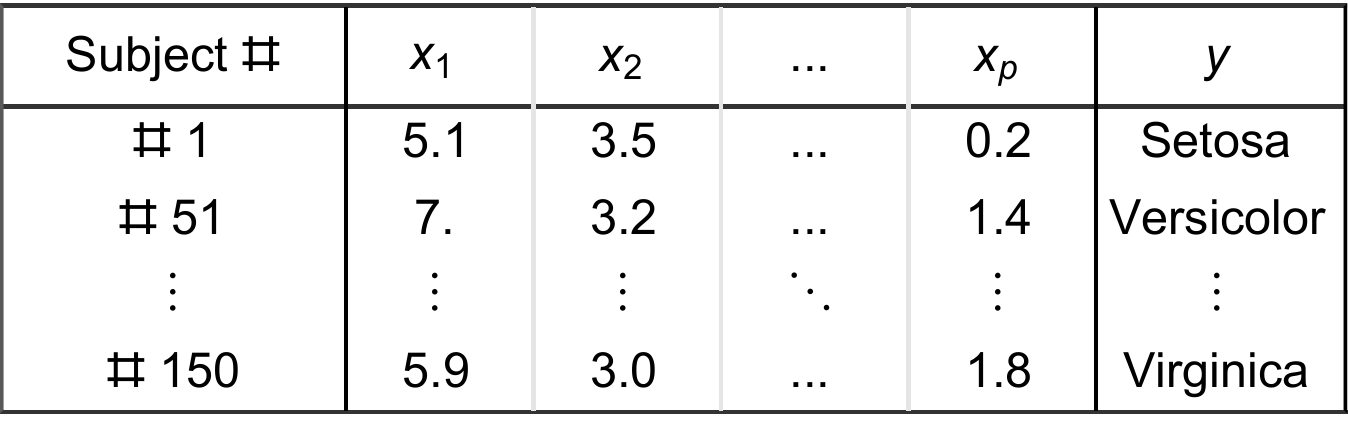}
\par\end{centering}
\caption{\label{fig Exemple Data-frame}Un exemple de data-frame $\mathcal{D}$ }
\end{figure}

Les données nécessaires à l'ajustement d'un classifieur doivent être
structurées dans ce qu'on appelle un data-frame, que l'on note aussi
$\mathcal{D}$. Un data-drame est un tableau rectangulaire ($n$ lignes
et ($p+1)$ colonnes) comme celui de la Figure \ref{fig Exemple Data-frame}.
Chaque ligne de $\mathcal{D}$ regroupe l'ensemble des $(p+1)$ valeurs
observées sur un des ($n$) individus faisant partie de $\mathcal{D}$.
Une ligne est à son tour structurée sous la forme :

\begin{align*}
(x_{1},...x_{p}\left|y\right.) & =(\boldsymbol{x}\left|y)\right.,
\end{align*}
où $\boldsymbol{x}=(x_{1},...,x_{p})$ regroupe les quantités $x_{i}$
appelées les caractéristiques, \emph{features}, \emph{insigths}, variables
de procuration, etc., qui permettent la classification. La quantité
$y$ est la cible, l'outcome ou la variable à classifier. Le processus
de construction d'un classifieur consiste à découvrir les relations
mathématiques entre les $\boldsymbol{x}$ et $y$ de $\mathcal{D}$.
Cet objet mathématique est ici noté $\mathcal{P}$ (pour prédicteur,
qui est souvent synonyme de classifieur).

Quant à la nature de $(\boldsymbol{x}\left|y)\right.$, les caractéristiques
$x_{i}$ peuvent être de type \emph{continu} (pouvant prendre en principe
n'importe quelle valeur, comme la pression sanguine, le taux de cholestérol,
etc.), de type \emph{discret} (des entiers 0, 1, 2, ..., correspondant
par exemple au nombre d'occurrences d'un mot dans un spam), \emph{booléen
}ou\emph{ binaire} (vrai/faux, homme/femme), \emph{catégorique} \emph{multiclasse}
(A, B, AB et O pour les groupes sanguins) ou \emph{catégorique ordinale}
(S, M, L et XL pour des tailles de vêtement). Lorsque $\boldsymbol{x}$
est constitué de caractéristiques de types différents, on dit  qu'elles
sont \emph{mixtes}.

La cible $y$ est souvent de type binaire ou catégorique et les valeurs
possibles de $y$ sont alors appelées des \emph{labels}. Quand $y$
est de type continu, le classifieur s'appelle un \emph{régresseur}.

\subsection{Prétraitement de $\mathcal{D}$}

La première partie de la phase d'ajustement d'un classifieur $\mathcal{P}$
consiste à effectuer un prétraitement de $\mathcal{D}$. Au-delà du
besoin évident de valider les données et de corriger les erreurs,
ce prétraitement peut affecter $y$ (on peut décider de regrouper
quelques labels d'un $y$ catégorique ou de catégoriser un $y$ continu),
mais il concerne surtout $\boldsymbol{x}.$ On peut estimer souhaitable
de catégoriser un $x_{i}$ continu, ou de le standardiser (ramener
sur une échelle centrée sur 0 par exemple), ou de combiner plusieurs
caractéristiques (p. ex. $x_{i}\times x_{j}$ ), bref d'effectuer
des opérations de toutes sortes ayant pour but de mettre en valeur
l'information discriminante des $\boldsymbol{x}$ de $\mathcal{D}$
pour préparer la suite de la phase d'ajustement du classifieur $\mathcal{P}$.
Ces opérations découlent des éléments cognitifs de $\mathcal{C}$
mais aussi des particularités du classifieur (par exemple, paramétrique
vs non paramétrique) que l'on entend utiliser. Ce prétraitement requiert
une intervention humaine et nécessite une expertise qui peut être
lourde. Un classifieur peu exigeant qui ``tout seul'' sait se débrouiller
avec les $\boldsymbol{x}$ qui lui sont donnés est donc avantageux.
Cette propriété, appelée invariance (aux transformations des données)
dans le jargon statistique, facilite considérablement l'ajustement
et l'interprétation des résultats du classifieur.

Signalons que dans la suite, on suppose que les données $\mathcal{D}$
sont de bonne qualités. En particulier, on suppose que les mesures
des $\boldsymbol{x}$ ne contiennent pas de valeurs atypiques, de
saisies erronées, etc. On suppose aussi que la détermination du label
de la cible $y$ est correcte. Dans de nombreuses applications, les
données ont été recueillies dans un but autre que celui de faire de
la classification. De ce fait, le niveau de rigueur dans la mesure
et l'enregistrement des données, que l'on retrouve dans les études
expérimentales en sciences, peut être déficient; par exemple $\mathcal{D}$
peut contenir des individus atypiques ne ``rentrant pas dans les
cases'' auxquels on a néanmoins attribué un label fantaisiste, etc.
La qualité d'un classifieur, particulièrement ceux de la variété non
paramétrique, dépend de la qualité des données selon le principe GI-GO
(= \emph{garbage in/garbage out}). Aussi abondant soit-il, si $\mathcal{D}$
est erroné, mais néanmoins utilisé, le classifieur peut mener à de
fausses conclusions et à des décisions préjudiciables. On peut retenir
que la qualité des données vaut bien plus de celle du classifieur.

\subsection{Le compromis biais/variance}

\begin{figure}
\begin{centering}
\includegraphics[scale=0.6]{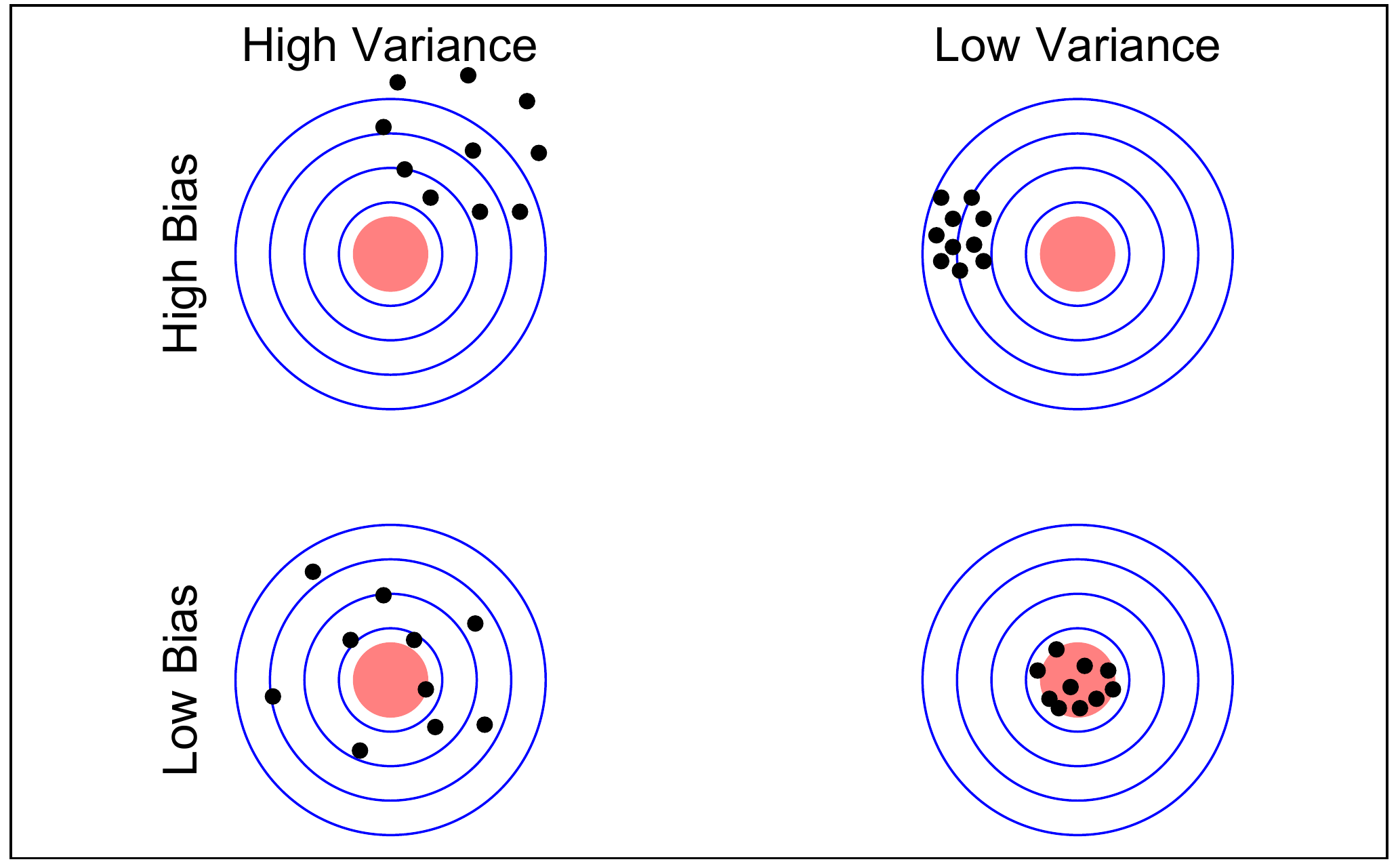}
\par\end{centering}
\caption{\label{fig:bias/variance compromise}Le compromis entre le biais et
la variance}

\end{figure}

Un prétraitement particulièrement important est la sélection des caractéristiques
$x_{i}$ à utiliser autant lors de l'ajustement du classifieur que
de son exploitation. En effet, il n'existe pas de ``grand livre de
la nature'' listant, pour chaque problème, les caractéristiques $x_{i}$
nécessaires à l'obtention de bonnes prédictions de la cible $y$.
Par ailleurs, certaines caractéristiques ont un lien causal avec la
cible, et donc un impact direct sur celle-ci, alors que d'autres ont
plutôt un lien corrélationnel avec la cible (impact plus diffus) ou
par d'autres caractéristiques qui, elles, ont un lien causal (impact
indirect). La théorie statistique montre qu'en général omettre des
$x_{i}$ liés d'une façon ou d'une autre à la cible peut créer un
\emph{biais} : le classifieur ``vise'' en moyenne à côté de la réalité.
Mais la théorie montre aussi qu'ajouter des caractéristiques inutiles
augmente la \emph{variance,} ou la variabilité, du classifieur. C'est
le fameux \emph{compromis biais/variance} qu'il faut, d'une façon
ou d'une autre, arbitrer. La Figure \ref{fig:bias/variance compromise}
explique le problème : un biais important rend le classifieur peu
utile, une variance importante le rend peu fiable. 

Lorsqu'une caractéristique importante a été ``oubliée'' au moment
de la constitution de $\mathcal{D}$, il est en général difficile
de rattraper après coup cette omission. C'est pourquoi en pratique
on tente d'éviter le problème en incluant dans $\mathcal{D}$ toutes
les caractéristiques imaginables qui pourraient impacter la cible,
à la manière d'un chalut, quitte à ajouter à la phase de prétraitement
une étape de filtration des $x_{i}$ inutiles. Mais cette nouvelle
phase, encore là, demande une supervision experte qui dépasse le niveau
du présent travail et fera l'objet d'une communication ultérieure.

\subsection{Ajustement et exploitation du classifieur}

Une fois le prétraitement effectué, on peut passer à la phase \emph{d'ajustement}
du classifieur. Cet ajustement, qui consiste à sélectionner la valeur
des hyperparamètres adaptés au problème considéré, fait intervenir
$\mathcal{C}$ et $\mathcal{D}$ (maintenant prétraité) d'une façon
complexe et spécifique à chacun des classifieurs et demande une solide
expérience de la part de l'utilisateur. La Section \ref{sec:Comparaison-de-certains}
présente une courte description des hyperparamètres devant être ajustés
pour certains classifieurs parmi les plus populaires et des problèmes
associés à leur détermination. Évidemment, un classifieur qui est
facile à ajuster est préférable. Une fois l'ajustement complété, on
dispose d'un classifieur, que l'on note $\hat{\mathcal{P}}$, et qui
est prêt à être utilisé en mode \emph{exploitation} : une valeur $\boldsymbol{x}^{*}$
devient accessible pour un individu appelé \emph{prospect} et pour
lequel on veut savoir la valeur inconnue $y^{*}$ de son label. La
valeur prédite $\hat{y}^{*}$ de $y^{*}$ est obtenue en injectant
$\boldsymbol{x}^{*}$ dans le classifieur $\hat{\mathcal{P}}$ (qui,
on le rappelle, est une fonction mathématique) :

\begin{alignat*}{1}
\hat{y}^{*} & =\hat{\mathcal{P}}(x^{*}).
\end{alignat*}

On poursuit l'exploitation tant qu'il y a des prospects à prédire.
En pratique, il se produit souvent avec les données mesurées sur des
humains un phénomène appelé la \emph{dérive des populations}. Cette
dérive fait qu'après un certain temps, les prospects ne sont plus
comme les individus de $\mathcal{D}$ (dans un contexte marketing,
les clients s'adaptent à une réalité économique qui fluctue) et la
valeur réelle $y^{*}$ de la cible dérive de la valeur prédite $\hat{y}^{*}$.
Ce phénomène apparaît notamment, mais pas seulement, dans les cas
où $\mathcal{C}$ est pauvre et $\mathcal{\hat{P}}$ s'appuie lourdement
sur $\mathcal{D}$. Un classifieur doit donc être périodiquement révisé
; $\mathcal{C}$ est réactualisé, des données fraîches $\mathcal{D}$
sont obtenues et le cycle \emph{ajustement}/\emph{exploitation} reprend.
On signale à la Section \ref{subsec:Les-crit=0000E8res-souhaitables}
une méthodologie statistique permettant de retarder le moment de la
révision. Mais éventuellement, cette révision s'impose et il importe
d'avoir en place un processus de monitorage pour la déclencher. Ce
dernier problème dépasse cependant le cadre du présent travail.

\section{Critères de qualité d'un classifieur\label{sec:Section 3 Crit=0000E8res-de-qualit=0000E9}}

Les sections précédentes ont montré quelques-unes des difficultés
liées à l'ajustement et à l'exploitation de classifieurs. Ayant en
tête les objectifs du présent travail, il en ressort qu'à moins d'être
un spécialiste de l'utilisation d'un classifieur, il est raisonnable
de rechercher, pour un usage généraliste, un classifieur qui :

\medskip{}

\begin{itemize}
\item \emph{demande un minimum d'intervention humaine} (en phase d'ajustement)
: l'absence de feuille de route dans la phase de prétraitement des
données est inconfortable pour un utilisateur qui se trouve face à
de nombreux choix, sans véritable protocole pour les piloter. De même,
la sélection des hyperparamètres et l'arbitrage du compromis biais/variance
sont souvent consternants. Le classifieur idéal devrait être peu exigeant
ou présenter une certaine insensibilité à ces choix, ou incorporer
des stratégies automatisant ces choix de façon fiable.
\item \emph{ait un faible taux d'erreur} : comme le fait remarquer Gayler
(2006), la plupart des classifieurs, lorsque bien ajustés, donnent
des taux d'erreur comparables, selon l'une ou l'autre des métriques
de l'\emph{accuracy} d'un classifieur (Hand, 2012). Encore faut-il
savoir choisir les hyperparamètres avec doigté, ce qui peut demander
une intervention humaine experte.
\item \emph{ait une ``operational relevance''} (en phase d'exploitation)
: il est rare que la seule classification de prospects suffise à l'utilisateur.
Dans un contexte où l'analytique prédictif est généralement suivi
d'analytique prescriptif, il importe que les sorties du classifieur
puissent s'insérer facilement dans un écosystème de méthodes et d'algorithmes
permettant la prise de décisions complexes. Ceci impose des contraintes
supplémentaires au classifieur.
\end{itemize}
\medskip{}

La synthèse de ces trois points fait ressortir des concordances, mais
aussi des antagonismes, et dans un tel contexte seul un compromis
peut être dégagé. Pour procéder de façon méthodique à la recherche
de ce compromis, nous allons constituer une liste des critères que
devrait rencontrer un classifieur généraliste ``idéal''. Plus loin,
nous attribuerons des scores selon ces critères à un certain nombre
de classifieurs populaires afin de dégager le meilleur compromis parmi
eux. 

La constitution de cette liste fait ressortir que certains critères
sont \emph{déterminants} : on peut difficilement recommander, pour
un usage généraliste, un classifieur qui ne les rencontre pas. Ces
critères sont listés à la Section \ref{subsec:Les-crit=0000E8res-fondamentaux}.
D'autres critères sont \emph{souhaitables} (Section \ref{subsec:Les-crit=0000E8res-souhaitables}).
Enfin, d'autres critères relèvent de considérations plus légères.
Ils n'ont pas acquis pour le moment suffisamment d'importance pour
justifier un score. Ils sont listés à la Section \ref{subsec:Les-crit=0000E8res-l=0000E9gers}.

Notons que cette liste de critères n'a pas la prétention de l'exhaustivité
et s'appuie largement sur des considérations subjectives tirées de
l'expérience de l'auteur. Par ailleurs, le souci que nous avons d'inscrire
notre démarche dans une perspective où l'analytique prédictif sera
suivi d'une phase analytique prescriptif, fait en sorte que cette
liste englobe et généralise certaines des listes, plus sommaires,
que l'on peut retrouver dans la littérature (Hans, 2002 (Sec. 2),
Lantz, 2015 ; EliteDataScience, 2018). Aussi, dans certaines niches
d'applications, l'importance de certains de ces critères pourrait
être atténuée, voire éliminée, alors que d'autres pourraient apparaître
ou gagner en importance. Néanmoins, il est à espérer que, dans ces
contextes particuliers, la présente liste constitue un point de départ
utile pour l'évaluation des classifieurs.

Notons par ailleurs que nous n'allons pas ici présenter, discuter
ou commenter les propriétés mathématiques des classifieurs, comme
leur vitesse de convergence et les majorations de type ``oracle''.
Nous nous concentrons sur des propriétés pratiques.

\subsection{Les critères déterminants\label{subsec:Les-crit=0000E8res-fondamentaux}}

Les critères déterminants d'un bon classifieur généraliste concernent
la phase d'ajustement du classifieur et portent sur sa précision (critère
3.1.1), sa bonne utilisation de $\mathcal{C}$ et $\mathcal{D}$ (3.1.2
et 3.1.3) et son automatisation (3.1.4 et 3.1.5). 

\medskip{}

\begin{enumerate}
\item[(3.1.1)] \emph{Le taux d'erreur}. La valeur $\hat{y}^{*}$ donnée par un classifieur
doit être proche de la vérité $y$ (inconnue). La façon naturelle
de mesurer la qualité d'un classifieur est de quantifier l'écart entre
$\hat{y}^{*}$ et $y$ : plus l'écart est faible, meilleur est le
classifieur. Mais, outre la difficulté de choisir une métrique du
taux d'erreur pertinente (p. ex. dans la liste de Hand, 2012), cette
propriété du classifieur est \emph{problem-specific} : sa compréhension
demande la synthèse de connaissances acquises concernant le contexte
pratique du problème, de considérations de statistique théorique,
de simulations en laboratoire et d'expériences de terrain. Il est
déterminant que l'utilisateur puisse s'appuyer sur une expérience
accumulée assurant que le classifieur offre généralement un faible
taux d'erreur (p. ex. peu d'exemples où il s'échoue lamentablement).
\item[(3.1.2)] \emph{L'exploitation des caractéristiques de type mixte.} Un classifieur
qui n'exploite pas toute l'information contenue dans $\mathcal{D}$
est à proscrire. Notamment, un classifieur qui ne gère pas (ou mal,
ou après des ``contorsions'') des caractéristiques ou une cible
d'un type donné n'exploite pas bien une information qui peut avoir
été coûteuse à obtenir.
\item[(3.1.3)] \emph{L'incorporation des informations de $\mathcal{C}$.} De la
même façon, un classifieur qui ne sait pas utiliser l'information
cognitive disponible est déficient. Par exemple, en marketing, on
a en général une bonne idée du coût du \emph{churn}. Incorporer cette
information dans la construction du classifieur permet d'optimiser
les actions et de trouver la stratégie engendrant les moindres coûts.
\item[(3.1.4)] \emph{Demander la sélection d'un minimum d'hyperparamètres}. On a
vu l'importance du bon choix des hyperparamètres. En gros, il y a
deux sortes d'hyperparamètres. Pour ceux qui sont des nombres, il
existe parfois des méthodes automatisant leur sélection (voir par
exemple le package caret de R) . Mais ces méthodes ne fournissent
que de premières approximations. D'autres hyperparamètres sont des
fonctions et, dans ces cas, seule l'expérience de l'utilisateur peut
aider à leur sélection. Il est donc fondamental qu'un classifieur
minimise le nombre et la complexité des hyperparamètres nécessaires
à son ajustement.
\item[(3.1.5)] \emph{Ne pas requérir d'optimisation problématique}. Mathématiquement,
de nombreux classifieurs sont la solution d'un problème d'optimisation.
Les algorithmes d'optimisation mathématique sont parfois délicats
à utiliser (optimums locaux, points de selle, matrices singulières,
etc.) ce qui donne des classifieurs difficiles à ajuster, instables
ou carrément faux. Un audit de leur déroulement est nécessaire, mais
peut demander l'intervention d'un expert. Il est important qu'un classifieur
soit peu exigeant à ce chapitre.
\end{enumerate}
\medskip{}

\subsection{Les critères souhaitables\label{subsec:Les-crit=0000E8res-souhaitables}}

Les critères souhaitables concernent l'\emph{operational relevance}
de Gayler (2006) et la facilité d'insertion des sorties du classifieur
dans l'écosystème d'algorithmes de l'analytique prescriptif. Ils se
rapportent à la simplicité d'exploitation et d'interprétation du classifieur
(3.2.1, 3.2.2, 3.2.3 et 3.2.9), à l'extraction d'information \emph{problem-specific}
pour l'analytique prescriptif (3.2.4 et 3.2.5) et à la confiance que
l'on peut conférer à ses sorties (3.2.6, 3.2.7 et 3.2.8). 

\medskip{}

\begin{enumerate}
\item[(3.2.1)] \emph{La capacité de gérer naturellement une cible mixte.} La version
originelle de certains classifieurs concerne des cibles d'un type
particulier (p. ex. binaire). La plupart ont éventuellement été adaptés
aux autres types, au prix parfois de ``contorsions''. Un classifieur
s'adaptant naturellement à tous les types de cibles est souhaitable
pour l'interprétation et la prise de décision.
\item[(3.2.2)] \emph{Invariance}. Pour des raisons d'interprétabilité, il est souhaitable
que le classifieur présente une certaine invariance (ou stabilité,
ou insensibilité) aux transformations et d'opérations effectuées sur
$\mathcal{D}$ lors du prétraitement. 
\item[(3.2.3)] \emph{Calibrage des scores}. La plupart des classifieurs retournent
des scores. Ces scores sont liés aux probabilités d'appartenance d'un
prospect à un label $y^{*}$ de la cible si sa caractéristique est
$\boldsymbol{x}^{*}$, ce que l'on note $\mathbb{P}[y^{*}\mid\boldsymbol{x}^{*}].$
L'exploitation de telles probabilités, dans un contexte marketing
notamment, permet d'estimer le coût moyen de certaines opérations.
Mais la conversion des scores en probabilités demande souvent un calibrage
et il est souhaitable que ce calibrage puisse être effectué facilement. 
\item[(3.2.4)] \emph{Estimation de la loi de probabilité de la cible}. Ce point
est lié au précédent. Au lieu de la seule valeur $\mathbb{P}[y^{*}\mid\boldsymbol{x}^{*}]$,
certaines opérations d'analytique prescriptive ont besoin de toute
la loi de probabilité de la cible, soit les valeurs de $\mathbb{P}[y^{*}\mid\boldsymbol{x}^{*}]$
pour toutes les valeurs $y^{*}$ possibles. \foreignlanguage{italian}{Cette}
loi de probabilité est plus informative, mais aussi plus complexe
à obtenir. Il est donc souhaitable de pouvoir l'estimer avec précision.
De façon similaire, certains problèmes, où l'on doit classifier $m$
prospects dont les caractéristiques sont $\boldsymbol{x}_{1}^{*},...,\boldsymbol{x}_{m}^{*}$
, nécessitent l'estimation de l'ordre des $m$ probabilités $\mathbb{P}[y^{*}\mid\boldsymbol{x}_{i}^{*}]$,.
Ce type de problème se rencontre par exemple en marketing, lorsqu'il
est nécessaire de concentrer les actions sur les prospects les plus
susceptibles du comportement désiré. Encore là, il est souhaitable
que le classifieur puisse produire une bonne estimation de cet ordre.
Mais ceci doit être tempéré à la lumière des travaux de Friedman (1997)
qui a montré théoriquement l'antagonisme entre la précision de l'estimation
de ces probabilités et le taux d'erreur du classifieur.
\item[(3.2.5)] \emph{Régularisa}t\emph{ion}. La régularisation est une opération
statistique qui force le classifieur à ne pas trop coller aux données.
Elle est un des outils permettant de retarder le problème de la dérive
des populations (Friedman, 2006). Bien que cet outil soit délicat
à manipuler et nécessite souvent une intervention experte, il est
souhaitable que le classifieur puisse l'intégrer. 
\item[(3.2.6)] \emph{Robustesse}. Malgré tous les soins mis à la constitution d'un
$\mathcal{D}$ de qualité, il peut néanmoins s'y glisser des valeurs
atypiques, erronées, etc. Il est souhaitable que le classifieur soit
peu sensible à de telles erreurs. Une approche consiste à rendre moins
sensibles, ou plus robustes, les fonctions optimisées par l'algorithme
du classifieur. Il est souhaitable que le classifieur puisse éventuellement
implémenter cette approche.
\item[(3.2.7)] \emph{Sélection de caractéristiques}. Il a été mentionné à la Section
\ref{subsec:Les donn=0000E9es} que $\mathcal{D}$ provient souvent
d'une ``campagne de chalutage'' où $p$ est très grand. Or l'utilisation
de caractéristiques inutiles augmente la variabilité des sorties du
classifieur. Il importe donc de savoir identifier les caractéristiques
contenant une réelle information discriminante. Plusieurs approches
sont possibles. Certaines s'appuient sur des mesures d'association.
D'autres opèrent à l'interne de l'algorithme du classifieur en produisant
une mesure de l'importance de chacune des caractéristiques, additionnée
ou non d'une procédure pas-à-pas de sélection. La combinaison des
deux approches est corroborative. Il est souhaitable qu'un classifieur
puisse produire de telles mesures d'importance.
\item[(3.2.8)] \emph{Traitement des caractéristiques manquantes}. Il arrive parfois
que des caractéristiques soient manquantes, de façon \emph{missing
at random,} pour quelques individus de $\mathcal{D}$ ou parmi les
prospects. En phase d'ajustement, certains algorithmes éliminent ces
cas (p. ex. retranchent des lignes de $\mathcal{D})$, ce qui mène
à une perte d'information. En phase d'exploitation, le classifieur
peut ne pas produire de prédiction pour un prospect. Il est souhaitable
qu'un classifieur arrive à contourner ce problème.
\end{enumerate}
\begin{quote}
\medskip{}
\end{quote}

\subsection{Autres critères\label{subsec:Les-crit=0000E8res-l=0000E9gers}}

Les critères de cette section concernent des préoccupations émergentes
qui pourraient prendre de l'importance dans le futur. D'autres sont
des points qui assurent un certain confort dans l'utilisation du classifieur.

\medskip{}

\begin{enumerate}
\item[(3.3.1)] \emph{Ne pas requérir trop de puissance informatique}. Au-delà des
ressources informatiques nécessaires à la constitution, à l'entretien
et au prétraitement de $\mathcal{D}$, il est confortable que les
processus d'ajustement et d'exploitation du classifieur ne soient
pas trop gourmands en temps de calcul, proportionnellement à $n$
et à $p$, ou puissent exploiter des avancées informatiques architecturales,
comme la parallélisation des tâches ou l'usage de GPU. Voir aussi
Hand (2012).
\item[(3.3.2)] \emph{Résistance aux attaques malicieuses.} On a découvert récemment
que certains classifieurs peuvent être ``attaqués'', c'est-à-dire
amenés à produire la classification qu'un adversaire souhaite (Goodfellow
\emph{et al.,} 2017). La gravité de telles failles apparaît, par exemple,
quand des caméras de surveillance doivent prendre les numéros d'immatriculation
de véhicules pour détecter les excès de vitesse. Pour le moment, cette
utilisation malicieuse est une curiosité limitée à la reconnaissance
d'images, mais de telles attaques pourraient déborder à d'autres contextes. 
\item[(3.3.3)] \emph{Transformation d'un problème de clustering en problème de classification}.
Jusqu'ici, on a discuté exclusivement du cas où les valeurs des labels
de la cible $y$ sont connues précisément au moment de l'ajustement
du classifieur. Or la détermination de ces labels est souvent un travail
coûteux, difficile ou demandant une intervention humaine, parfois
sous-traitée à des agents à la fiabilité inconnue (p. ex. via \emph{Amazon
Mechanical Turk}). Dans les cas où la cible est manquante ou trop
imprécisément connue, on pourrait se contenter d'un regroupement des
prospects en segments. Une méthodologie pour ce faire est le \emph{clustering},
mais elle nécessite qu'une distance entre les prospects soit déterminée.
Certains classifieurs produisent naturellement une telle distance.
\item[(3.3.4)] \emph{Transparence}. En phase d'analytique prescriptif, des décisions
seront prises, ou des actions entreprises, en fonction des sorties
du classifieur. La législation de certains pays exige, dans le cas
notamment de \emph{credit scoring}, que le fonctionnement du classifieur
puisse être transparent. De façon plus générale, la transparence des
principes et concepts sur lesquels s'appuie le classifieur est une
source de confiance en ses résultats. 
\item[(3.3.5)] \emph{Attrait intellectuel du classifieur}. Il n'est pas inutile
d'être séduit par les concepts qui sous-tendent un classifieur. Par
exemple, les réseaux de neurones artificiels (Section \ref{subsec:Reseaux-de-neurones})
sont censés imiter le comportement des neurones cérébrauxs. Ces liens
entre les mathématiques et la biologie sont attrayants pour ceux qui
sont impressionnés par les solutions que la nature a développées pour
assurer la survie des espèces, et qui confèrent des vertus aux méthodes
qui s'en inspirent. De façon générale, l'attrait intellectuel est
un élément motivateur poussant à acquérir une expertise du classifieur,
ce qui ne peut être que bénéfique.
\end{enumerate}
\medskip{}

\section{Quelques classifieurs populaires \label{sec:Comparaison-de-certains}}

Armés des critères de la section précédente, nous allons maintenant
comparer six classifieurs basés sur des concepts différents. Ceux-ci
font partie des classifieurs les plus populaires, selon l'appréciation
de l'auteur et certaines compilations que l'on rencontre dans la littérature.
En particulier, en décembre 2006 s'est tenue la \emph{IEEE International
Conference on Data Mining }(ICDM). Lors de cette conférence, les dix
algorithmes de \emph{data-mining} les plus utilisés ont été identifiés.
Parmi ceux-ci, on note les classifieurs C4.5, SVM, AdaBoost, kNN,
Naive Bayes et CART (Wu et al., 2008). Deux autres articles similaires
plus récents (Dezire, 2018; EliteDataScience, 2018) listent en gros
les mêmes classifieurs, avec quelques ajouts reflétant les fluctuations
de popularité. Dans la suite, nous nous limitons aux six classifieurs
suivants :

\medskip{}

\begin{itemize}
\item Bayes naïf (\textbf{Bn}), (classifieur),
\item Régression logistique (\textbf{RLog}), (classifieur),
\item Séparateur à Vaste Marge ou \emph{Support Vector Machine} (\textbf{SVM}),
(classifieur / régresseur),
\item Forêt aléatoire (\textbf{RF-CART}), (classifieur/régresseur),
\item $k$ plus proches voisins (\textbf{kNN}), (classifieur/régresseur),
\item Réseau de neurones artificiels (\textbf{RNA}), (classifieur/régresseur).
\end{itemize}
\medskip{}

Certains classifieurs, notamment les plus anciens, ont été éliminés
de cette liste par l'un ou l'autre des critères déterminants (discrimination
linéaire de Fisher (1936), quadratique, etc.). D'autres n'ont pas
résisté à l'usure du temps, ou leur utilisation est restreinte à certaines
niches ; pour la tâche de notation qui suit, il importe d'avoir une
expérience du comportement des classifieurs sur une palette plutôt
large de problèmes. D'autres encore, en particulier les arbres CART,
CI, C4.5, etc., sont des éléments de la classe ensembliste des forêts
d'arbres, de laquelle on a choisi le représentant ``Forêt aléatoire
\textbf{RF-CART}''. Nous sommes conscients que la plupart des experts
data scientists affectionnent une petite famille de classifieurs,
qu'ils utilisent selon leurs besoins. Le fait qu'un de leurs classifieurs
de prédilection ne se retrouve pas dans notre collection n'est pas
une appréciation de notre part de sa mauvaise qualité. Au besoin,
ils peuvent ajouter ce dernier aux tableaux de scores de la Section
\ref{sec:Section 5 Tableaux-des-scores} et reconstruire les comparaisons. 

Nous allons maintenant brièvement présenter ces six classifieurs.
Chacun d'entre eux est appuyé par une énorme littérature qui explore
ses propriétés, ses particularités et ses adaptations à différents
problèmes. Il est impossible de rendre justice à cette littérature
dans les courtes descriptions qui suivent. On peut espérer que ces
résumés ne donnent pas une image trop étriquée de leurs comportements. 

\subsection{Bayes naïf (Bn)}

Il s'agit d'un classifieur très simple dont l'ajustement demande peu
d'intervention humaine et aucune maximisation. Il suppose l'indépendance
entre les caractéristiques, ce qui ne tient en réalité jamais. Mais
on a découvert (Hand \& Yu, 2001) que même si cette hypothèse ne tient
pas, le classifieur \textbf{Bn} fonctionne étonnamment bien pour une
large palette de problèmes. Il gère les valeurs manquantes sans difficulté.
Il fonctionne bien avec un $\mathcal{D}$ limité ($n$ petit) et incorpore
facilement les éléments de $\mathcal{C}$. Il est souvent le premier
classifieur auquel on pense pour la classification de textes. Par
contre, son utilisation dans la phase d'analytique prescriptif subséquente
est suspecte du fait que les scores du classifieur ne sont valides
que dans ce contexte d'indépendance et risquent d'être biaisés autrement.
Enfin, il n'est pas bien adapté au cas de cibles ou de caractéristiques
continues. Bref, un outil \emph{quick and dirty} pour la seule tâche
d'analytique prédictif.

\subsection{Régression logistique (RLog)\label{subsec:R=0000E9gression-logistique-(RLog)}}

La régression logistique\emph{ }est un classifieur ancien qui a été
développé pour des cibles binaires. Mais elle peut s'adapter au cas
catégorique au prix de certaines contorsions (en particulier les approches
\emph{un contre tous} ou \emph{un contre un}), ou en utilisant les
modèles de régression GLM. Étant donné sa très grande popularité,
notamment dans le domaine de la biostatistique et du \emph{credit
scoring}, il est difficile de ne pas la considérer dans la présente
étude comparative. C'est un classifieur paramétrique qui exige de
l'utilisateur le choix d'une structure qui se détermine souvent en
pratique par essai et erreur, ou par commodité lorsque $\mathcal{C}$
est pauvre. Mais le taux d'erreur dépend de ce choix arbitraire. 

En revanche, lorsque $\mathcal{C}$ est riche et que le choix de la
structure est fait de façon éclairée, comme en biostatistique ou en
\emph{credit scoring}, l'aspect paramétrique est un avantage certain
qui permet de comprendre et d'interpréter les sorties de ce classifieur.
La \textbf{RLog} est facilement régularisable et certains experts
arrivent même à modéliser directement la dérive de la population,
de façon à prolonger la vie utile du classifieur (en \emph{credit
scoring}, cette durée est de 3 à 5 années selon Gayler, 2006). De
plus, un expert peut créer, de la combinaison des caractéristiques
existantes, de nouvelles caractéristiques (p. ex. par ACP) pour former
des modèles qui prennent une teinte semi-paramétrique.

Les procédures d'optimisation pour l'estimation des paramètres sont
classiques, mais posent parfois problème en raison de colinéarités
entre des caractéristiques. La gestion des caractéristiques mixtes
se fait aussi d'une façon un peu folklorique (\emph{dummy variable}),
ce qui peut poser des problèmes quand $p$ est grand. L'incorporation
d'information de $\mathcal{C}$ est possible. Les scores ont une interprétation
probabiliste directe. Bref, un bon classifieur (mais pas un régresseur)
avec un fort potentiel en analytique prescriptif pour autant que la
structure paramétrique ait été correctement trouvée. Heureusement,
des outils diagnostiques existent, permettant de détecter ce problème
et d'offrir des pistes de solution. Il vaut mieux cependant qu'un
expert se tienne à proximité de l'utilisateur. 

\subsection{Séparateur à vaste marge (SVM)}

Les \textbf{SVM} ont été introduits par Vapnik (1995) et ont connu
un développement fulgurant en raison, d'une part, de la beauté mathématique
de leur concept et, d'autre part, de leurs excellentes performances
en matière de taux d'erreur dans un grand nombre de problèmes comme
la reconnaissance de formes, de séquences génomiques, de spams, etc.
Ils ont été conçus originellement comme classifieur pour cible binaire,
mais ont été étendus au cas catégorique (au prix des mêmes contorsions
que la \textbf{RLog}) et au cas de régresseur au prix d'une complexité
mathématique peut-être rebutante. Par ailleurs, une sélection judicieuse
des hyperparamètres (il y en a plusieurs à choisir, notamment les
noyaux des \textbf{SVM}) permet de traiter le cas des caractéristiques
mixtes et de nombreux cas non standards, comme la classification de
texte. Mais cette sélection implique, en phase d'ajustement, une ingénierie
de noyaux complexe qu'il est risqué d'effectuer sans intervention
experte puisque ces noyaux sont des fonctions. À ce jour, il ne semble
pas exister d'outils diagnostiques indiquant si la sélection est bien
ou mal adaptée. L'intégration des éléments cognitifs de $\mathcal{C}$,
le traitement de caractéristiques manquantes et l'étude de l'importance
des variables ne sont pas encore bien développés. Pour ces problèmes,
des solutions \emph{ad hoc} existent, mais elles demandent aussi une
supervision humaine et leur fiabilité n'est pas encore bien établie.
Il est possible (par l'algorithme de Platt) de calibrer leurs scores
à des fins d'analytique prescriptif. Enfin, la construction d'un classifieur
\textbf{SVM} exige une optimisation qui, sans être trop périlleuse,
n'en est pas moins exigeante sur le plan informatique, dès lors que
le problème devient plus complexe que la simple classification binaire.

\subsection{Forêt Aléatoire (RF, \emph{random forest})\label{subsec:Random-Forest-(RF)}}

Les arbres de classification CART ont été inventés par Breiman \emph{et
al.} (1984) et leur version ensembliste, la forêt aléatoire \textbf{RF-CART},
par Breiman (2001). Tous les classifieurs de cette famille sont maintenant
très populaires et l'expérience accumulée sur leur comportement est
vaste.

Les arbres de classification CART présentent de nombreux avantages.
Leur taux d'erreur est faible, et cette appréciation s'appuie autant
sur des simulations en laboratoire que sur d'innombrables applications
de terrain. Ils sont non paramétriques, robustes, simples à construire
et ne demandent quasiment pas de prétraitement. Ils acceptent les
caractéristiques de toutes natures et les cibles de tous types. Ils
procèdent automatiquement à la sélection des hyperparamètres et gèrent
de façon transparente les caractéristiques manquantes. Aucune optimisation
complexe n'est requise et ils incorporent naturellement des informations
de $\mathcal{C}$. Enfin, ils produisent en sortie une structure d'arbre
facile à interpréter et à communiquer. Les scores ont une interprétation
probabiliste et peuvent être facilement calibrés (algorithme de Platt
ou régression isotonique). Ils produisent des mesures de l'importance
des variables. Ils ne demandent quasiment aucune intervention humaine.
Ils ont quelques difficultés à gérer des caractéristiques catégorielles
où le nombre de catégories est grand, mais ce problème n'est pas spécifique
aux arbres.

Une\emph{ }forêt aléatoire CART (\textbf{RF-CART}) est constituée
d'un ensemble d'arbres CART. De ce fait, elle possède la plupart des
avantages de ces arbres et en diminue le taux d'erreur. D'ailleurs,
les différentes versions des forêts aléatoires se retrouvent souvent
sur le podium dans les compétitions de classifieurs. Ce gain de précision
se fait principalement au détriment de l'interprétation, car une forêt
aléatoire est de type ``boîte noire''. Elle gère mal les données
manquantes. Elle nécessite l'introduction de deux hyperparamètres
scalaires, mais l'algorithme est assez insensible à leur sélection,
pour autant que celui-ci soit raisonnable et proche des valeurs par
défaut suggérées, qu'on peut donc en général utiliser en confiance
(Liaw \& Wiener, 2002). Elle est naturellement parallélisable.

Au vu des nombreux avantages des arbres et des forêts aléatoires,
plusieurs variantes, corrigeant l'un ou l'autre des faiblesses de
la version CART dans certaines niches d'applications, ont vu le jour
(C4.5, C5.0, CI, CHAID, etc.). Nous ne développons pas ces variantes
dans ce travail et nous nous contentons de considérer les \textbf{RF
CART} comme représentant de cette famille.

\subsection{k plus proches voisins (kNN)}

La méthode des k plus proches voisins est une méthode de prédiction
simple et ancienne, remontant au moins à Cover \& Hart (1967). C'est
une méthode \emph{instance-based} (c.-à-d. sans modèle) donc non paramétrique,
s'adaptant à des cibles de tous types (classification et régression)
et n'exigeant pas d'optimisation. Elle peut produire de bons taux
d'erreur.

Un problème avec cette approche est la sélection des hyperparamètres.
Un premier hyperparamètre à déterminer est la valeur $k$ du nombre
de voisins, laquelle dépend de la position locale des caractéristiques,
et il existe des méthodes permettant d'automatiser ce choix. Cet hyperparamètre
joue aussi un rôle en tant que régulateur du classifieur. Mais il
faut de plus choisir la métrique définissant le terme ``proche'',
ce qui, quand les caractéristiques sont mixtes, demande une ingénierie
de métriques de complexité comparable à la sélection de noyaux pour
les \textbf{SVM} et requiert une supervision experte. \textbf{kNN}
est beaucoup plus simple à utiliser quand toutes les caractéristiques
sont du type continu, car alors une seule métrique peut être utilisée
(métrique euclidienne, choisie par commodité généralement). Elle peut
s'accommoder des caractéristiques de type catégoriel, mais le travail
est plus laborieux. Il en va de même des caractéristiques manquantes.
Des poids aussi peuvent être associés aux voisins, mais encore là
un noyau (hyperparamètre) doit être sélectionné.

\textbf{kNN} demande aussi une phase de prétraitement lourde pour
éviter que les échelles des caractéristiques ne viennent confondre
leur pouvoir de discrimination. De plus, la méthode se dégrade sévèrement
en présence de caractéristiques inutiles, de sorte qu'il faut faire
un tri important à la phase de prétraitement. En particulier, on recommande
souvent une phase de réduction de dimension des caractéristiques par
l'ACP. \textbf{kNN} peut produire des scores, mais il semble qu'on
n'ait pas encore exploré la possibilité de les calibrer. Aussi, comme
la méthode ne construit pas de modèles, elle est une boîte noire qui
ne permet pas de comprendre et d'interpréter les liens entre les caractéristiques
et la cible. 

Bref, \textbf{kNN} peut être un bon classifieur généraliste dans les
problèmes simples et dans les cas où un sérieux prétraitement a été
fait. Autrement, son ajustement demande un pilotage expert de haut
niveau.

\subsection{Réseaux de neurones artificiels (RNA)\label{subsec:Reseaux-de-neurones}}

Les réseaux de neurones artificiels (\textbf{RNA}) remontent aux travaux
de McCulloch \& Pitts (1943). Leurs principes reposent sur un modèle
simplifié du comportement des véritables neurones cérébraux, dont
dérive une partie de leur attrait. Cependant, le cerveau humain contient
environ 85 milliards de neurones et celui d'une blatte, environ un
million. Les RN modernes, n'ayant en général guère plus que quelques
milliers de neurones, ne sont donc qu'une version limitée des cerveaux
que l'on rencontre dans la nature. 

Après quelques hauts et bas, les \textbf{RNA} ont considérablement
gagné en popularité dans les années 1990 et depuis l'arrivée des méthodologies
du \emph{deep learning} (circa 2010) et les succès des algorithmes
de jeux (DeepBlue pour les échecs, AlphaGo pour le jeu go, etc.),
de classification d'images, de sons, de textes, etc., ils sont l'objet
d'une forte médiatisation. Mais leurs indéniables succès dans certains
problèmes pointus ne se transposent pas nécessairement à d'autres
domaines, notamment les domaines touchant les comportements humains,
où les données sont moins riches que pour les jeux (on estime que
AlphaGo a dû ingurgiter un $\mathcal{D}$ équivalent à quelques millénaires
d'expériences du jeu go (Clark, 2017, p. 38) pour réussir son exploit.
L'explication de cette instabilité des performances fait d'ailleurs
l'objet de recherches intenses sur le plan théorique dans le but de
la comprendre et de la corriger. Pour le moment, ces recherches pointent
vers la complexité trop grande des réseaux qui sont utilisés. (Zhang
\emph{et al}, 2017).

Les \textbf{RNA} les plus simples (les perceptrons) coïncident avec
la \textbf{RLog} et partagent donc les forces et les faiblesses de
cette méthode. Leurs extensions à des réseaux plus complexes (les
réseaux à couches cachées du \emph{deep learning} et autres raffinements)
dévient des modèles statistiques et forment une nouvelle classe de
classifieurs dont les informaticiens sont les indéniables champions.
Ils demandent la détermination de la topologie du réseau de neurones
artificiel et pour le moment, ce choix se fait par essai et erreur.
Une certaine latitude est cependant autorisée, différentes topologies
pouvant donner des prédictions comparables. Ils comportent de très
nombreux paramètres et hyperparamètres. Ils souffrent souvent de problèmes
d'optimisation importants nécessitant une supervision humaine vigilante,
demandent des ressources informatiques très lourdes et un prétraitement
conséquent. D'ailleurs, de nombreuses recherches récentes du \emph{Google
Brain Team} (Dean, 2018) visent à automatiser le processus d'ajustement
d'un \textbf{RNA}. Par ailleurs, les \textbf{RNA} ont tendance à surajuster
les données et le contrôle du compromis biais/variance (et d'éventuelles
régularisations) est délicat. Ils sont conçus pour utiliser principalement
les données de $\mathcal{D}$ et l'incorporation d'éléments de $\mathcal{C}$
n'est pas toujours simple. Ils produisent des scores mais il ne semble
pas qu'on se soit beaucoup penché sur leur calibration. Leur traitement
des données manquantes et atypiques (auquel on réfère par le terme
\emph{fault tolerance,} Clark, 2017) est primitif et ils n'offrent
qu'une faible possibilité d'étude de l'importance des variables. Ils
constituent une boîte noire qu'il est quasiment sans espoir de déchiffrer.
Leur précision peut cependant être remarquable dans certaines niches
particulières. Ils fonctionnent pour des problèmes de classification
et de régression. Bref des outils pouvant être très efficaces, mais
demandant une grande supervision et pas encore bien adaptés à la phase
subséquente d'analytique prescriptif.

\section{Tableaux des scores\label{sec:Section 5 Tableaux-des-scores}}

Hand (2012) déplore l'absence de métrique pour comparer les classifieurs
selon des critères \emph{problem-specific} et il encourage la recherche
sur ce problème. Dans l'intervalle, nous allons ici attribuer de façon
subjective des scores aux différents classifieurs de la Section \ref{sec:Comparaison-de-certains}
selon les critères de la Section \ref{sec:Section 3 Crit=0000E8res-de-qualit=0000E9}.
L'échelle de ces scores va de 1 (note minimale) à 5 (note maximale).
Dans l'absolu, aucun classifieur ne peut recevoir le score 5 car nos
critères ne définissent pas de normes à rencontrer parfaitement définies.
De la même façon et toujours dans l'absolu, aucun classifieur ne peut
recevoir le score 1 car, étant tous très populaires, aucun n'a de
déficiences graves. On a donc reporté des scores relatifs, où la note
5 est attribuée au(x) classifieur(s) satisfaisant le mieux le critère
considéré, selon l'appréciation de l'auteur, et de même pour le score
1, pour ceux qui s'en éloignent le plus. Le but principal de l'exercice
est de ranger les classifieurs et ces scores relatifs permettent d'accentuer
les écarts entre des classifieurs qui ont tous des vertus. On a ensuite
calculé la moyenne des scores, pour former les notes de chacun des
six classifieurs. 

Une part de subjectivité intervient forcément dans l'attribution de
ces scores et nous espérons ne pas avoir été systématiquement biaisés
ou injustes dans leur attribution. Néanmoins, nous pensons que l'ordre
des classifieurs est assez juste et stable. Ces tableaux sont donc
le reflet d'une certaine réalité et d'une expérience certaine, mais
ne peuvent avoir de sens que sur une courte fenêtre de temps. Ces
tableaux devront évoluer avec l'accumulation des connaissances sur
les performances et les fluctuations de la popularité des divers classifieurs. 

Pour les critères fondamentaux, les résultats, scores et notes, se
retrouvent dans la Table \ref{tab:Crit=0000E8res 3.1}. On voit que
les \textbf{RF} obtiennent la meilleure note, suivis d'un cluster
formé de \textbf{RLog}, \textbf{kNN} et enfin de \textbf{Bn, RNA}
et \textbf{SVM}. Ces deux derniers perdent des points essentiellement
en raison des difficultés associées à leur ajustement alors que pour
\textbf{Bn}, c'est la trop grande simplicité du classifieur qui pose
problème.

\begin{table}
\begin{centering}
\begin{tabular}{|l|c|c|c|c|c|c|}
\hline 
Critères (3.1.\emph{x}) & \textbf{Bn} & \textbf{RLog} & \textbf{SVM} & \textbf{RF} & \textbf{kNN} & \textbf{RNA}\tabularnewline
\hline 
\hline 
1 - Taux d'erreur & 1 & 4 & 5 & 5 & 2 & 5\tabularnewline
\hline 
2 - Carac. Mixte & 1 & 3 & 3 & 5 & 3 & 3\tabularnewline
\hline 
3 - Info. de $\mathcal{C}$ & 2 & 4 & 3 & 5 & 1 & 3\tabularnewline
\hline 
4 - Hyperpara. & 5 & 4 & 1 & 5 & 1 & 2\tabularnewline
\hline 
5 - Optimisation & 5 & 4 & 1 & 5 & 5 & 1\tabularnewline
\hline 
\hline 
Note /5 & 2.8 & 3.6 & 2.6 & 5 & 3.2 & 2.8\tabularnewline
\hline 
\end{tabular}
\par\end{centering}
\caption{\label{tab:Crit=0000E8res 3.1}Évaluation des classifieurs selon les
critères déterminants. Les scores sont relatifs, allant de 1 à 5.}
\end{table}

Concernant les critères souhaitables, la Table \ref{tabCrit=0000E8res 3.2}
présente les résultats en suivant la même démarche. Encore ici, les
\textbf{RF} l'emportent, suivis de la paire formée de \textbf{RLog
}et \textbf{SVM}. Enfin \textbf{RNA}, \textbf{Bn} et \textbf{kNN}
ferment la marche, le caractère \emph{instance-based} de la \textbf{kNN}
la pénalisant fortement par rapport à la compétition.

Globalement, on peut conclure qu'en regard de nos deux listes de critères,
le meilleur classifieur est la forêt aléatoire (\textbf{RF}). Le classifieur
\textbf{RLog} est aussi intéressant, mais limité aux cibles catégorielles.
Les autres classifieurs demandent une expertise de plus haut niveau
lors de la phase d'ajustement et ne positionnent pas toujours favorablement
l'utilisateur pour la phase d'analytique prescriptif. Ils peuvent
être excellents dans des niches particulières, mais leur utilisation
dans un contexte généraliste est plus délicate à ce stade de leur
développement. 

\begin{table}
\begin{centering}
\begin{tabular}{|l|c|c|c|c|c|c|}
\hline 
Critères (3.2.\emph{x}) & \textbf{Bn} & \textbf{RLog} & \textbf{SVM} & \textbf{RF} & \textbf{kNN} & \textbf{RNA}\tabularnewline
\hline 
\hline 
1 - Cible mixte & 2 & 1 & 4 & 5 & 4 & 5\tabularnewline
\hline 
2 - Invariance & 2 & 3 & 3 & 5 & 1 & 3\tabularnewline
\hline 
3 - Calibrage Score & 2 & 5 & 4 & 5 & 1 & 3\tabularnewline
\hline 
4 - Loi de prob. cible & 2 & 5 & 3 & 4 & 1 & 3\tabularnewline
\hline 
5 - Régularisation & 1 & 5 & 4 & 5 & 3 & 4\tabularnewline
\hline 
6 - Robustesse & 3 & 5 & 4 & 4 & 1 & 2\tabularnewline
\hline 
7- Sélec. caractéristiques & 2 & 5 & 3 & 5 & 1 & 1\tabularnewline
\hline 
8 - Valeurs manquantes & 5 & 2 & 2 & 3 & 1 & 1\tabularnewline
\hline 
\hline 
Note /5 & 2.38 & 3.88 & 3.38 & 4.5 & 1.63 & 2.75\tabularnewline
\hline 
\end{tabular}
\par\end{centering}
\caption{\label{tabCrit=0000E8res 3.2}Évaluation des classifieurs selon les
critères souhaitables. Les scores sont relatifs, allant de 1 à 5.}
\end{table}

\section{Conclusion}

La prédiction du futur est en général un problème difficile et il
est naturel de se focaliser sur la précision, ou le taux d'erreur.
Ceci explique les nombreux travaux en \emph{machine learning }qui
visent à développer des classifieurs maximisant cette métrique. Ces
travaux ont rencontré un grand succès : les classifieurs modernes,
notamment ceux présentés à la Section \ref{sec:Comparaison-de-certains},
sont très efficaces. Si les données possèdent un pouvoir de discrimination,
ces classifieurs arriveront probablement à l'exploiter au mieux. 

Mais en vertu des théorèmes de l'introduction, ces prédictions ne
peuvent jamais être parfaites et l'on doit se contenter de probabilités
de leur réalisation. Ceci modifie complètement le paradigme de pensée
: dès lors que des risques apparaissent, il devient raisonnable d'accepter
de payer un prix afin de se prémunir d'éventuelles désastres. Dans
ce paradigme, à la base du \emph{statistical learning}, une stratégie
légèrement sous-optimale, pouvant occasionner au pire des inconforts,
pourrait être largement préférable à une stratégie optimale aux conséquences
catastrophiques. 

C'est dans ce contexte que nous avons présenté une liste de critères
complétant le taux d'erreur en y ajoutant des points se rapportant
à l'utilisation qui sera faite de ces classifieurs, soit l'\emph{operational
relevance }de Gayler (2006). Nous avons considéré l'amont comme l'aval
de la phase d'analytique prédictif proprement dite. En amont, le classifieur
doit demander le prétraitement le plus léger possible. En aval, il
doit fournir des résultats qui serviront ensuite d'entrées dans l'écosystème
d'algorithmes de l'analytique prescriptive. De façon transversale,
il ne doit pas constituer une charge de travail trop lourde ou nécessiter
une expertise humaine trop pointue.

La \textbf{RF-CART }et ses variantes ressortent de ce travail comme
étant les meilleurs classifieurs parmi ceux que nous avons étudiés.
Ces classifieurs font partie de la classe ensembliste et la recherche
en ce domaine, et notamment sur les \textbf{RF}, est extrêmement active,
tant sur le plan théorique (Scornet, 2015) qu'heuristique (Davies
\& Ghahramani, 2014). Par ailleurs, une étude informelle (Argerich,
2016) indique qu'au moment d'écrire ces lignes, les \textbf{RF} auraient
la préférence dans de nombreuses entreprises américaines. Comme cette
préférence est probablement le résultat du cheminement empirique de
leurs data scientists, il apparaît donc que nos conclusions peuvent
servir de justificatif pour soutenir ce cheminement. Ceci conforte
dans l'importance et la valeur du présent travail. 

Comme suite, nous discuterons dans un article à venir des différentes
variétés de forêts d'arbres aléatoires, pour en dégager les avantages
et inconvénients dans les mêmes perspectives que celles du présent
travail.

\bigskip{}

\textbf{Remerciements} : L'auteur remercie ses collègues de l'Université
de Montpellier pour des discussions qui ont contribué à l'élaboration
de ce travail. L'auteur assume cependant l'entière responsabilité
des recommandations et suggestions faites.

\bigskip{}


\begin{thebibliography}{10}
\bibitem{key-1}Argerich, L. (2015) : What are the most important
machine learning algorithms? What are the most commonly applied algorithms
when attacking a problem? \url{https://www.quora.com/What-are-the-most-important-machine-learning-algorithms-What-are-the-most-/commonly-applied-algorithms-when-attacking-a-problem}.

\bibitem{key-50}Bhaowal, M. (2018) : Square off : Machine learning
libraries. \url{https://www.oreilly.com/ideas/square-off-machine-learning-libraries?imm_mid=0fa832&cmp=em-data-na-na-newsltr_20180117}.

\bibitem{key-17}Boström, H. H. (2007) : Estimating class probabilities
in random forests. In \emph{Proc. of the International Conference
on Machine Learning and Applications,} 211\textendash 216.

\bibitem{key-44}Breiman, L., Friedman, J. H., Olshen, R. A., \& Stone,
C. J., (1984) \emph{Classification and regression trees}. Monterey,
CA, Wadsworth \& Brooks/Cole Advanced Books \& Software.

\bibitem{key-39}Breiman, L. (2001) : Random forests. \emph{Machine
Learning}, 45, 5\textendash 32.

\bibitem{key-14}Breiman, L. (2004) : Population theory for boosting
ensembles. \emph{Ann. Statist.}, 32, 1-11.

\bibitem{key-15}Briand, B., Ducharme, G.R, Parache,V., Mercat-Rommens,
C. (2009): A similarity measure to assess the stability of classification
trees. \emph{Computational Statistics and data analysis}, 53, p.1208-1217.

\bibitem{key-46}Clark, J. (2017) : \emph{Artifical Intelligence :
Teaching machine to think like people}. O'Reilly Media Inc., Sebastopol.

\bibitem{key-16}Coudray, A. (2017) : Trois applications concrètes
de l'intelligence artificielle en marketing. \emph{Journal du Net},
\url{http://www.journaldunet.com/ebusiness/expert/68014/3-applications-concretes-de-l-intelligence-artificielle-en-marketing.shtml}
.

\bibitem{key-18}Cover, T.M,. Hart, P.E. (1967). \textquotedbl{}Nearest
neighbor pattern classification\textquotedbl{}. \emph{IEEE Transactions
on Information Theory,} 13, 21\textendash 27.

\bibitem{key-48}Davies, A.,; Ghahramani, Z., (2014) : The Random
Forest Kernel and other kernels for big data from random partitions.
arXiv:1402.4293

\bibitem{key-20}Dean, J. (2018) : Looking back on 2017 (Part 1 of
2). \url{https://research.googleblog.com/2018/01/the-google-brain-team-looking-back-on.html?imm_mid=0fa832&cmp=em-data-na-na-newsltr_20180117.}.

\bibitem{key-21}Demsar J. (2006) : Statistical Comparisons of Classifiers
over Multiple Data Sets. \emph{The Journal of Machine Learning Research},
7, 1-30.

\bibitem{key-22}Dezure, (2018) : Top 10 Machine Learning Algorithms.
\url{https://www.dezyre.com/article/top-10-machine-learning-algorithms/202}.

\bibitem{key-23}Dietterich, T.G. (1998) : Approximate Statistical
Tests for Comparing Supervised Classification Learning Algorithms.
\emph{Neural Computation,} 10, 1895-1924. 

\bibitem{key-24}Dixit, A. (2017) : \emph{Ensemble machine learning}.
Packt Publishing Ltd., Birmingham.

\bibitem{key-25}EliteDataASciences (2018) : Modern Machine Learning
Algorithms: Strengths and Weaknesses. \url{https://elitedatascience.com/machine-learning-algorithms?imm_mid=0fa832&cmp=em-data-na-na-newsltr_20180117} 

\bibitem{key-27}Escoufier, Y., Fichet, B., Diday, E., Lebart,L.,
Hayashi, C., Ohsumi, N., Baha, Y. (1995) : Preface to ``\emph{Data
science and its application - La science des données et ses applications}''.
Escoufier et al. ed. , Academic Press, Tokyo. 

\bibitem{key-26}Fisher, R. A. (1936) : The use of multiple measurements
in taxonomic problems. \emph{Annals of Eugenics,} 7, 179\textendash 188.

\bibitem{key-5}Friedman, J.H. (1997) : On bias, variance, 0/1-loss,
and the curse of dimensionality. \emph{Data Mining Knowledge Discovery},
1, 55-77.

\bibitem{key-30}Friedman, J.H. (2006) : Comment : Classifier Technology
and the Illusion of Progress. \emph{Statistical Science}, 21, 15-18.

\bibitem{key-28}Gayler, R.W. (2006) : Comment : Classifier Technology
and the Illusion of Progress. \emph{Statistical Science}, 21, 19-23.

\bibitem{key-31}Goodfellow, I.. Shlens. J., Szegedy, C. (2017) :
Explaining and Harnessing Adversarial Examples. \url{ https://research.google.com/pubs/pub43405.htm}.

\bibitem{key-19}Hand, D.J., Keming, Y. (2001) : Idiot's Bayes\textemdash Not
So Stupid After All? \emph{International Statistical Review}, 69,
385-398.

\bibitem{key-32}Hand, D.J. (2006) : Classifier Technology and the
Illusion of Progress. \emph{Statistical Science}, 21, 1-14.

\bibitem{key-3}Hand, D.J. (2012) : Assessing the performance of classification
methods. \emph{International Statistical Review}, 80, 400-414.

\bibitem{key-2}Hand, D.J., Anagnostopoulos, C. (2013) : When is the
area under the receiver operating characteristic curve an appropriate
measure of classifier performance. \emph{Pattern Recognition Letters},
34, 492-495.

\bibitem{key-1}Jamain, A., Hand, D.J. (2008) : Mining supervised
classification performance : A meta-analytic investigation. Journal
of Classification, 25, 87-112.

\bibitem{key-34}Kuhn, M. (2008) : Building Predictive models in R
using the caret package. Journal of Statistical Software, 28, Issue
5. \url{http://www.jstatsoft.org/}.

\bibitem{key-38}Lantz, B. (2015) : \emph{Machine learning with R.
2nd edition}. Packt Publishing Ltd., Birmingham.

\bibitem{key-45}Liaw, A., Wiener, M. (2002) : Classification and
regression by random forest. R News, vol 2/3, 18-22.

\bibitem{key-33}Lim, T.J., Low, W.Y \& SHih, Y.S. (2000) : A comparison
of prediction accuracy, complexity and training time of thirty-three
old ans new classification algorithms. Machine Learning, 40(3), 203-228.

\bibitem{key-35}Lustig, I., Dietrich, B., Johnson, C., Dziekan, C.
(2010) : The Analytics Journey. \url{http://www.analytics-magazine.org/november-december-2010/54-the-analytics-journey}.

\bibitem{key-36}McCulloch,W.S., Pitts, W. (1943) : A logical calculus
of the ideas immanent in nervous activity. \emph{Bulletin of Mathematical
Biophysics,} 5:115-133.

\bibitem{key-37}Ohsumi, N. (2000) : From data analysis to data science.
In \emph{Proceedings of the 7th conference of the International Federation
of Classification Societies} (IFCS-2000), Kiers et al. Ed. Springer.
ISBN-13:978-3-540-67521-1.

\bibitem{key-47}Scornet, E. (2015). Random forests and kernel methods.
arXiv:1502.03836\,

\bibitem{key-6}Vapnik V., (1995) : \emph{The nature of statistical
learning theory}, Springer-Verlag, New York, NY, USA, 1995.

\bibitem{key-7}Wolpert DH., (1996) : The Lack of A Priori Distinctions
Between Learning Algorithms. \emph{Neural Computation}, 8, 1341-1390.

\bibitem{key-8}Wu, X. Kumar, V., Quinlan, J.R.,Ghosh. J., Yang, Q.,
Motoda, H., McLachlan, G. J., Ng, A., Liu, B., Yu, P.S., Zhou,Z.,
Steinbach,M., Hand, D.J., Steinberg, D. (2008) : Top 10 algorithms
in data mining. \emph{Knowledge and Information System}, 14, 1.37.

\bibitem{key-1}Zhang, C., Bengio, A., Hardt, M., Recht, B., Vinyals,
O. (2017) : Understanding deep Learning requires rethinking generalization.
ICLT 2017, arXiv:1611.03530v2 {[}cs.LG{]} for this version)
\end{thebibliography}
\end{document}